\documentclass[runningheads]{llncs}
\usepackage[T1]{fontenc}
\usepackage{graphicx}
\usepackage{cite}
\usepackage{amsmath,amssymb,amsfonts}
\usepackage{xcolor}
\usepackage{xurl}

\begin{document}
\title{Forecasting Cryptocurrency Prices using Contextual ES-adRNN with Exogenous Variables\thanks{\footnotesize G.D. thanks prof. W.K. Härdle for his guidance on cryptocurrencies. G.D. and P.P. were partially supported by grant 020/RID/2018/19 "Regional Initiative of Excellence" from the Polish Minister of Science and Higher Education, 2019-23.}}
\titlerunning{Forecasting Cryptocurrency Prices using cES-adRNN}
\author{Slawek Smyl\inst{1}\orcidID{0000-0003-2548-6695}
\and Grzegorz Dudek \inst{2}\orcidID{0000-0002-2285-0327} \and
Paweł Pełka\inst{2}\orcidID{0000-0002-2609-811X}}
\authorrunning{S. Smyl et al.}
\institute{\email{slawek.smyl@gmail.com} \and
Electrical Engineering Faculty, Czestochowa University of Technology, Poland\\
\email{\{grzegorz.dudek,pawel.pelka\}@pcz.pl}}
\maketitle              % typeset the header of the contribution
\begin{abstract}

In this paper, we introduce a new approach to multivariate forecasting cryptocurrency prices using a hybrid contextual model combining exponential smoothing (ES) and recurrent neural network (RNN). The model consists of two tracks: the context track and the main track. The context track provides additional information to the main track, extracted from representative series. This information as well as information extracted from exogenous variables is dynamically adjusted to the individual series forecasted by the main track.  
The RNN stacked architecture with hierarchical dilations, incorporating recently developed attentive dilated recurrent cells, allows the model to capture short and long-term dependencies across time series and dynamically weight input information. The model generates both point daily forecasts and predictive intervals for one-day, one-week and four-week horizons. We apply our model to forecast prices of 15 cryptocurrencies based on 17 input variables and compare its performance with that of comparative models, including both statistical and ML ones.

\end{abstract}
\section{Introduction}

Forecasting cryptocurrency prices is a very difficult task due to several reasons. Firstly, the cryptocurrency market is highly volatile and unstable, with prices fluctuating rapidly and unpredictably \cite{Giu20}. This is because the market is not regulated, and there is no central authority that controls the supply and demand of cryptocurrencies. Unlike traditional assets such as stocks, cryptocurrencies lack a fundamental value. This means that their value is not directly linked to any underlying assets or earnings. Instead, their value is primarily determined by market demand and supply dynamics, making their price highly volatile and subject to sudden fluctuations. Secondly, cryptocurrencies are relatively new, and there is a lack of historical data, making it difficult to identify trends and patterns. Thirdly, the market is influenced by various factors \cite{Sov18, Wal19}, including government regulations (related to the legality of cryptocurrencies, taxation policies, and anti-money laundering laws), global economic conditions, and the emergence of new cryptocurrencies, which can make it challenging to predict price movements accurately. Moreover, market sentiment and speculation can have a significant impact on cryptocurrency prices. Cryptocurrencies are still a relatively new and emerging asset class, and as such, they are more susceptible to hype, media coverage, and market sentiment. This can cause price bubbles and crashes, making it challenging to forecast their prices accurately.
Finally, the cryptocurrency market operates 24/7, and there is no closing or opening bell, which means that prices can change at any time, making it challenging to keep up with the latest developments and adjust forecasts accordingly.

To capture the influence of external factors on the cryptocurrency price, the forecasting models use different inputs, which can be categorized as follows \cite{Gra23,Mud20,Wal19, Bou19, Ahm22, Kra20, Saa20}:

\begin{enumerate}
\item Economic indicators. Cryptocurrency prices may be influenced by macroeconomic indicators such as GDP, inflation, and interest rates.
\item Trading volume. The volume of cryptocurrency trades can indicate market demand, which can affect price movements.
\item Technical indicators. Technical analysis tools such as moving averages, Bollinger bands, and RSI can help identify trends and patterns in price movements.
\item News sentiment. The sentiment analysis of news articles and social media posts related to cryptocurrencies can provide insight into the market's mood and direction.
\item Network activity. The activity on the blockchain network, such as the number of transactions and the hash rate, can reflect the popularity and adoption of a particular cryptocurrency.
\item Regulatory developments. Government regulations and policies related to cryptocurrencies can significantly impact their prices.
\end{enumerate}

Multivariate forecasting models used for cryptocurrency prices can be categorized into three groups: classical statistical or econometrics models, computational intelligence or machine learning (ML) models, and hybrid models \cite{Khe21}. One popular representative of the classical statistical models is the Vector Autoregression (VAR) model \cite{Hot18, Giu19}. This model uses a system of equations to estimate the relationship between multiple variables and can capture the dynamic interdependence between them. However, it assumes linear relationships between variables, which may not always hold in the highly complex and nonlinear cryptocurrency market. Therefore, it may not be suitable for accurately predicting cryptocurrency prices in all scenarios.

ML models have several advantages over classical statistical models in forecasting cryptocurrency prices. One advantage is their ability to capture nonlinear relationships between variables. Additionally, ML models can handle large amounts of data, can effectively extract relevant features from it, and are often better than statistical models at identifying patterns and trends. They can also adapt to new information and adjust their forecasts accordingly. 
However, ML can be more computationally intensive and requires more data preprocessing compared to statistical models. Also, as a general rule, they require more data.

Some examples of ML models for forecasting cryptocurrency prices are as follows. 
A model proposed in \cite{Kim22} utilizes on-chain data as inputs, which are unique records listed on the blockchain that are inherent in cryptocurrencies. To ensure stable prediction performance in unseen price ranges, the model employs change point detection to segment the time series data. The model consists of multiple long short-term memory (LSTM) modules for on-chain variable groups and the attention mechanism.
The research described in \cite{Han22} uses a multivariate prediction approach and compare different types of recurrent neural networks (LSTM, Bi-LSTM, and GRU). Seven input variables are considered: date, open, high, low, close, adjusted close, and volume.
Paper \cite{Che23} aims to develop a prediction algorithm for Bitcoin prices using random forest regression and LSTM. The study also seeks to identify the variables that have an impact on Bitcoin prices. The analysis utilizes 47 explanatory variables, which are categorized into eight groups, including Bitcoin price variables, technical features of Bitcoin, other cryptocurrencies, commodities, market index, foreign exchange, public attention, and dummy variables of the week.
In \cite{Che20}, a set of high-dimension features including property and network, trading and market, attention and gold spot price was used for Bitcoin daily price prediction. The authors compared statistical models such as logistic regression and linear discriminant analysis with ML models including random forest, XGBoost, support vector machine and LSTM.
A comprehensive comparison of statistical and ML models for cryptocurrency price prediction can be found in \cite{Khe21}.

Our study introduces a hybrid model that merges statistical and ML methods, specifically exponential smoothing (ES) and recurrent neural networks (RNNs). This model, referred to as cES-adRNN, consists of two tracks designed to incorporate context information, effectively handle time series of exogenous variables and enhance the accuracy of forecasting. The proposed model offers several advantages over existing ML models:

\begin{itemize}
\item Unlike traditional ML models that often require data preprocessing to simplify the forecasting problem, our model can handle raw time series data without any preprocessing. It has built-in mechanisms for preprocessing time series dynamically on-the-fly.

\item While classical ML models require additional procedures to select relevant input information, our recurrent cells have an internal attention mechanism that dynamically weighs inputs, making the model more efficient.

\item Unlike non-recurrent ML models, RNN, which is a component of our model can model temporal relationships in sequential data. We apply dilated recurrent cells, which are fed by both recent and delayed states, to capture short-term, long-term, and seasonal dynamics. Moreover, our hierarchical RNN architecture extends the receptive fields in subsequent layers, enabling better modeling of long-term and seasonal relationships.

\item Most forecasting ML models produce only point forecasts. Our model is able to generate both point forecasts and predictive intervals in the same time.

\item While most forecasting ML models learn on a single time series, our model can learn from multiple similar series, leading to better generalization.

\end{itemize}

% https://onlinelibrary.wiley.com/doi/full/10.1002/isaf.1488
% https://link.springer.com/chapter/10.1007/978-3-030-49186-4_9
% https://www.sciencedirect.com/science/article/pii/S2214212620307535
% https://www.sciencedirect.com/science/article/pii/S037704271930398X

% https://link.springer.com/chapter/10.1007/978-981-15-5397-4_63
% https://link.springer.com/chapter/10.1007/978-3-030-72236-4_17
% https://link.springer.com/chapter/10.1007/978-3-030-72236-4_17

The main contributions of this study lie in the following three aspects:
\begin{enumerate}
\item We extend our cES-adRNN model proposed in \cite{Smy23} by introducing exogenous variables. This enables the model to predict time series based not only on the past values of the series (univariate case) but also on other, parallel time series which are correlated with the predicted series.

\item We modify cES-adRNN by introducing additional per-series parameters allowing to capture individual properties of the series of exogenous variables.

\item To validate the efficacy of our proposed approach, we conduct extensive experiments on 15 cryptocurrency datasets comprising 17 variables. Our experimental results show the high performance of the modified cES-adRNN model and its potential to forecast cryptocurrency prices more accurately than comparative models including both statistical and ML models.
\end{enumerate}

The remainder of this paper is organized as follows. In Section 2, we present the data and define a forecasting problem. Section 3 presents the proposed forecasting model. 
Section 4 reports our experimental results. Finally, we
conclude this work in Section 5.

\section{Data and Forecasting Problem}

The real-world data was collected from BRC Blockchain Research Center (\url{https://blockchain-research-center.com}; accessible for BRC members). The dataset BitInfoCharts provides a wide range of tools for accessing information on the cryptocurrency sphere. It includes market, blockchain operation, and social media data for 16 cryptocurrencies:
BTC, ETH, DOGE, ETC, XRP, LTC, BCH, ZEC, BSV, DASH, XMR, BTG, RDD, VTC, BLK, FTC. In this study, we omitted RDD from this list due to the short period of data (from 13 January 2021).

The dataset includes 19 variables (time series) for each cryptocurrency:
\begin{itemize} \small
\begin{multicols}{2}
\item \begin{verbatim} active_addresses_per_day \end{verbatim}
\item \begin{verbatim} avg_block_size_per_day \end{verbatim}
\item \begin{verbatim} avg_block_time_per_day \end{verbatim}
\item \begin{verbatim} avg_fee_to_reward \end{verbatim}
\item \begin{verbatim} avg_hashrate_per_day \end{verbatim}
\item \begin{verbatim} avg_mining_difficulty_per_day \end{verbatim}
\item \begin{verbatim} avg_price_per_day \end{verbatim}
\item \begin{verbatim} avg_transaction_fee_per_day \end{verbatim}
\item \begin{verbatim} avg_transaction_value_per_day \end{verbatim}
\item \begin{verbatim} google_trends \end{verbatim}
\item \begin{verbatim} market_capitalization \end{verbatim}
\item \begin{verbatim} med_transaction_fee_per_day \end{verbatim}
\item \begin{verbatim} med_transaction_value_per_day \end{verbatim}
\item \begin{verbatim} mining_profitability \end{verbatim}
\item \begin{verbatim} num_transactions_per_day \end{verbatim}
\item \begin{verbatim} num_tweets_per_day \end{verbatim}
\item \begin{verbatim} num_unique_addresses_per_day \end{verbatim}
\item \begin{verbatim} sent_coins \end{verbatim}
\item \begin{verbatim} top_100_to_all_coins \end{verbatim}
\end{multicols}
\end{itemize}

The data availability and quality differ from coin to coin, due to the issue time, crypto design, and the third party vendors. Due to incompleteness or poor quality, we excluded two variables, \texttt{google\_trends} and \texttt{num\_tweets\_per\_day}. 

Our goal is to predict  prices of cryptocurrencies (\texttt{avg\_price\_per\_day}) based on historical values of all variables. Thus, we predict a sequence of cryptocurrency prices of length $h$, $\{{z}_\tau\}_{\tau=M+1}^{M+h}$, given past observations of prices, $\{z_\tau\}_{\tau=1}^{M}$, and $N$ exogenous variables, $\{p^i_\tau\}_{\tau=1}^{M}$, $i=1, .., N$,  where $h$ is the forecast horizon and $M$ is the time series length (training part). In Fig. \ref{figP} the forecasted time series of cryptocurrency prices are shown. 

\begin{figure}[h]
	\centering
    \includegraphics[width=0.85\textwidth]{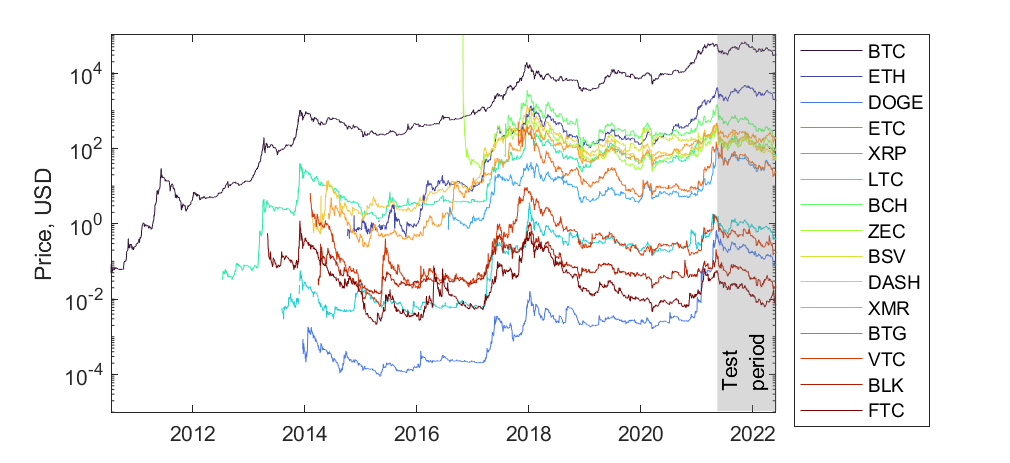}
    \caption{Prices of cryptocurrencies.} 
    \label{figP}
\end{figure}

We consider three forecast horizons: 1, 7 and 28 days ahead. For each horizon, a separate model is constructed. We define one-year test period from 1 June 2021 to 31 May 2022. Moving in the test period by one day, the model generates forecasts for the next $h$ days, using past data for training.         

\section{Model}

\begin{figure}[h]
	\centering
    \includegraphics[width=0.75\textwidth]{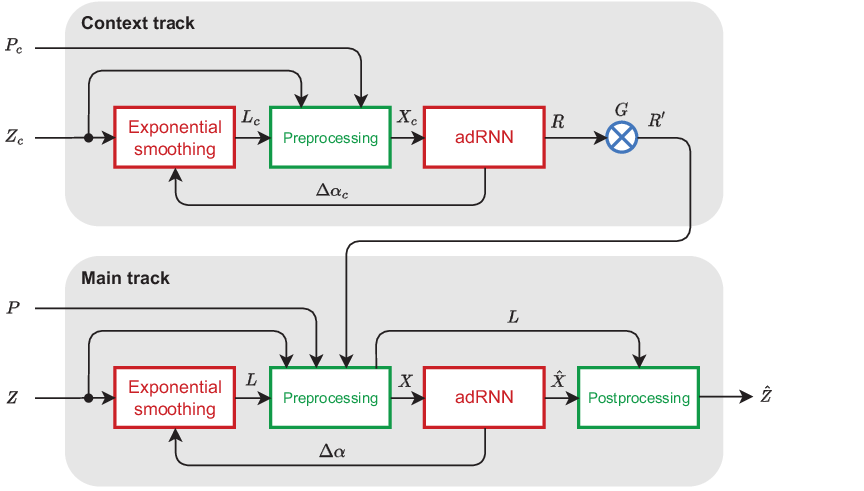}
    \caption{Block diagram of the forecasting model.} 
    \label{figBD}
\end{figure}

The model is a modified version of contextually enhanced ES-dRNN with dynamic attention (cES-adRNN) proposed in \cite{Smy23}. It combines exponential smoothing (ES) and recurrent neural network (RNN). The model is composed of two simultaneously trained tracks: context track and main track. The main track learns to generate point forecasts for horizon $h$ and also predictive intervals. The context track learns to generate additional inputs for the main track based on representative time series. The block diagram of the model is shown in Fig. \ref{figBD}. 

The model is trained in cross-learning mode (on all cryptocurrency data). Some components of the model are able to capture properties of individual time series, while others learn shared features of all series. Thus the data is exploited hierarchically, utilizing both local and global components to extract and synthesize information at both the individual series and collective dataset levels.

\subsection{Main Track}

Inputs to the main track are price time series ($Z$) and exogenous time series ($P$). 

\subsubsection{Exponential Smoothing Component} (ES) smooths price time series. It has a basic form given by the formula:
\begin{equation}
\begin{aligned}
l_{t,\tau}=\alpha_t z_\tau + (1-\alpha_t)l_{t,\tau-1} \\
\label{eq1}
\end{aligned}
\end{equation}
where $l_{t,\tau}$ is a level component, and $\alpha_t \in [0, 1]$ is a smoothing coefficient, both for recursive step $t$.

A dynamic variant of ES is utilized, wherein the smoothing coefficient is not constant, unlike in the standard version, but instead changes dynamically at each step $t$ to adapt to the current time series properties, as proposed in \cite{Smy22}. The adaptation of the smoothing coefficient is performed using correction $\Delta\alpha_t$, which is learned by RNN, as follows \cite{Smy22}:

\begin{equation}
\begin{aligned}
\alpha_{t+1} = \sigma(I\alpha + \Delta\alpha_t)\\
\label{eqab}
\end{aligned}
\end{equation}where $I\alpha$ represents the starting value of $\alpha$, and the sigmoid function $\sigma$ is employed to ensure that the coefficient remain within the range from 0 to 1. 

\subsubsection{Preprocessing Component} 
\begin{figure}[h]
	\centering
    \includegraphics[width=0.7\textwidth]{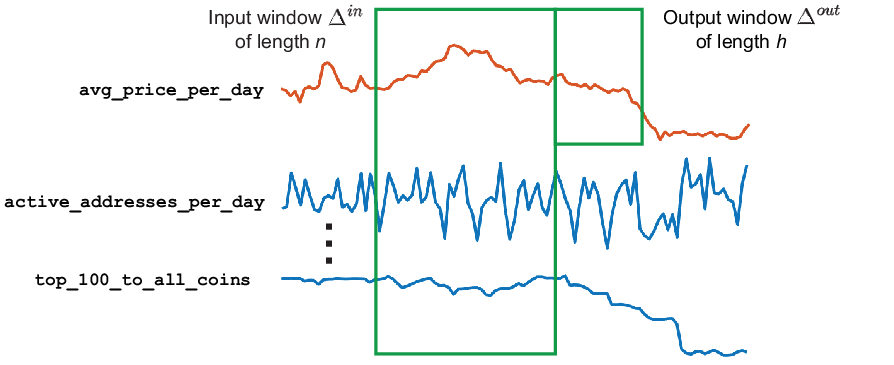}
\caption{Sliding windows for generating training data.} 
    \label{figSw}
\end{figure}

prepares input and output data for RNN training. Input data includes preprocessed sequences of the price and exogenous variables time series, i.e. sequences covered by the sliding input window of length $n$, $\Delta^{in}_t$, while output data includes preprocessed sequences of the price series covered by the sliding output window of length $h$, $\Delta^{out}_t$ (see Fig. \ref{figSw}).

To normalize both input and output price sequences, level component \eqref{eq1} extracted by ES is used. Input sequences are preprocessed as follows:
\begin{equation}
x_\tau^{in}=\log{\frac{z_\tau}{\hat{l}_{t}} }
\label{eqxt}
\end{equation}where $\tau \in \Delta^{in}_t$, and $\hat{l}_{t}$ is the level component predicted by ES for the last point in $\Delta^{in}_t$ in step $t$.

The use of the $\log$ function for squashing in \eqref{eqxt} is aimed at preventing outliers from interfering with the learning process.

The output sequences are normalized as follows:
\begin{equation}
x_\tau^{out}={\frac{z_\tau}{\hat{l}_{t}} }
\label{eqxt2}
\end{equation}where $\tau \in \Delta^{out}_t$.

The preprocessed input and output sequences are represented by vectors: $\textbf{x}_t^{in}=[x_\tau^{in}]_{\tau \in \Delta^{in}_t} \in \mathbb{R}^{n}$ and $\textbf{x}_t^{out}=[x_\tau^{out}]_{\tau \in \Delta^{out}_t} \in \mathbb{R}^{h}$, respectively. 
It should be noted that the input and output patterns have a dynamic nature and are modified in each training epoch due to the adaptation of the level component. This can be viewed as the learning of the optimal representation for RNN.

The input sequences of the $i$-th exogenous variable are normalized as follows:

\begin{equation}
x_\tau^{p_i}=\log_{10}\left({\frac{p^i_\tau}{\bar{p}^i}}+1\right)
\label{eqxpt}
\end{equation}
where $\tau \in \Delta^{in}_t$, and $\bar{p}^i$ is the average value of the $i$-th exogenous variable in the training period. 

The regressors take broad range of positive numbers and zero. 
Therefore the ratio $\frac{p^i_\tau}{\bar{p}^i}$ can be very small and occasionally zero. For this reason we add 1 in the formula above. The preprocessed input sequences of exogenous variables are represented by vectors $\textbf{x}_t^{p_i}=[x_\tau^{p_i}]_{\tau \in \Delta^{in}_t} \in \mathbb{R}^{n}$, $i=1, ..., N$.

To construct input patterns for RNN, vectors $\textbf{x}_t^{in}$ and $\textbf{x}_t^{p_i}, i=1, ..., N$ are concatenated and extended by two components: a local level of the price series, $\log_{10}(\hat{l}_t)$, and a modulated context vector produced by the context track, $\textbf{r}'_t$:

\begin{equation}
\textbf{x}_t^{in'}= [\textbf{x}_t^{in},\, \textbf{x}_t^{p_1}, ...,\, \textbf{x}_t^{p_N},\, \log_{10}(\hat{l}_t),\,  \textbf{r}'_t] 
\label{eqxp}
\end{equation}

\subsubsection{Recurrent Neural Network}
is trained on training samples $(\textbf{x}_t^{in'}, \textbf{x}_t^{out})$ generated continuously by shifting sliding windows $\Delta^{in}_t$ and $\Delta^{in}_t$ by one day.
 RNN produces four components: 
point forecast vector, $\hat{\textbf{x}}_t^{RNN}$, two quantile vectors defining a predictive interval (PI), $\hat{\underline{\textbf{x}}}_t^{RNN}$ and $\hat{\bar{\textbf{x}}}_t^{RNN}$, and correction for the smoothing coefficient, $\Delta\alpha_t$. 

RNN architecture is depicted in Fig. \ref{figP}. Initially, to reduce the dimensionality of the exogenous data, each $n$-dimensional vector $\textbf{x}_t^{p_i}$ is embedded into $d$-dimensional space ($d<n$) by a linear layer. Note that the embedding is trained as part of the model itself. 

After embedding, the exogenous vectors are concatenated and modulated by per-series parameters collected in modulation vectors $\textbf{p}^{(j)} \in \mathbb{R}^{dN}, j=1, ..., J$, where $J$ is the size of the main batch. Modulation vectors initialized as ones are used to element-wise multiply the exogenous vector:

\begin{equation}
\textbf{x}_t^{p'(j)} = \textbf{x}_t^{p(j)} \otimes \textbf{p}^{(j)}
\label{eqp}
\end{equation}

Modulation vectors $\textbf{p}^{(j)}$ are updated along with other learnable parameters of the model using the same optimization algorithm (stochastic gradient descent), with the the ultimate objective of minimizing the loss function.

\begin{figure}[h]
	\centering
    \includegraphics[width=0.7\textwidth]{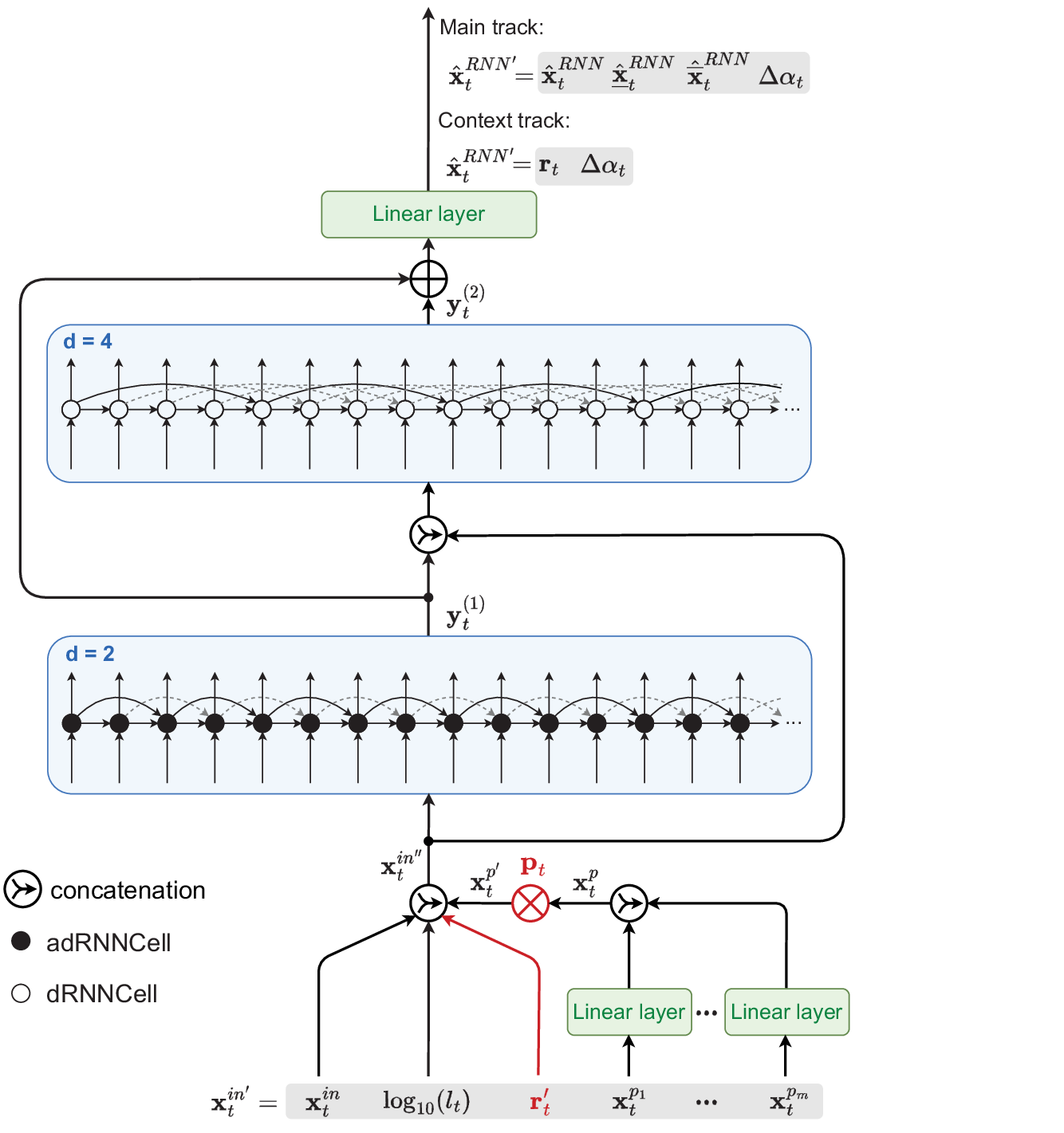}
\caption{Architecture of adRNN (elements in red do not exist in the context track).} 
    \label{figP}
\end{figure}

RNN utilizes two types of recurrent cells \cite{Smy22}: the dRNNCell, a dilated cell that is fed by both recent and delayed states, and the adRNNCell, a dilated cell with an attention mechanism. The latter combines two dRNNCells, with the first producing an attention vector which components are treated as weights for the inputs to the second, generating an output vector from which the forecast is constructed. Hence, the components of the input vector $\textbf{x}_t^{in'}$ can be strengthened or weakened by the attention vector depending on their predictive power. Note that the attention vector has a dynamic nature, as it is adapted to the current inputs at time $t$ (see \cite{Smy22} for details).

RNN is composed of two recurrent layers: first one contains adRNNCell dilated 2, while the second one contains dRNNCell dilated 4. 
The number of layers and dilations were determined through experimentation and may not be the best choice for other forecasting tasks. By stacking multiple layers with hierarchical dilations, more abstract features can be extracted in successive layers and a larger receptive field can be obtained, making it easier to learn the long-term dependencies of different scales. To facilitate the gradient optimization of the model, ResNet-style shortcuts are used between layers. It's worth noting that the input vector $\textbf{x}_t^{in'}$ is not only inputted into the first layer, but also into the second one by extending the output vector from the first layer.

\subsubsection{Postprocessing Component} 
converts outputs of RNN, i.e. point forecasts and PIs, into real values as follows:

\begin{equation}
\hat{z}_\tau=\exp{\left(\hat{x}_\tau^{RNN}\right)} {\hat{l}_{t}},\;\;\; \hat{\underline{z}}_\tau=\exp{\left(\hat{\underline{x}}_\tau^{RNN}\right)} {\hat{l}_{t}},\;\;\; \hat{\overline{z}}_\tau=\exp{\left(\hat{\overline{x}}_\tau^{RNN}\right)} {\hat{l}_{t}}
\label{eqzt}
\end{equation} where $\tau \in \Delta^{out}_t$, $\hat{\underline{z}}_\tau$ and $\hat{\overline{z}}_\tau$ are the lower and upper bounds of PI at time $\tau$.

\subsubsection{Loss Function}
enables the model to learn both point forecasts and PIs. It is defined as follows \cite{Smy21}:

\begin{equation}
L =
\rho(x, \hat{x}_{q^*}) + \gamma(
\rho (x, \hat{{x}}_{\underline{q}}) + 
\rho (x, \hat{{x}}_{\overline{q}}))
\label{eqlss}
\end{equation}
where 
$\rho(x, \hat{x}_q)=(x-\hat{x}_q)(q - \textbf{1}_{(x<\hat{x}_q)})$ is a pinball loss, $q \in (0, 1)$ is a quantile order, $x$ is an actual value, $\hat{x}_q$ is a forecasted value of $q$-th quantile of $x$, $q^*=0.5$ corresponds to the median, $\underline{q} \in (0,q^*)$ and $\overline{q} \in (q^*, 1)$ correspond to the lower and upper bound of PI, respectively, and $\gamma \geq 0$ is a control parameter.

The loss function \eqref{eqlss} enables both point forecasts and PIs to be optimized at the same time. The weight of each component can be adjusted by $\gamma$. The pinball loss has the advantage of reducing forecast bias by penalizing positive and negative deviations differently. This can be done by adjusting $q^*$ to be less than or greater than $0.5$. Similarly the bias in PI can be reduced. This concept is further explained in \cite{Smy20, Dud21}.

\subsection{Context Track}

A context track generates a dynamic context vector, which is based on the history of a representative group of series. This vector is adjusted to the series being forecasted and added as supplementary input to the main track RNN. This extends the per-series input data with information from the context series, thereby enhancing the accuracy of individual series forecasts. 

The context track is trained on the selected price time series ($Z_c$) and associated exogenous time series ($P_c$). In this study, Bitcoin price series along with its corresponding exogenous data are selected as the context series. This is because Bitcoin is the most popular, widely accepted and has the longest historical data among all cryptocurrencies. 

The context track components are: ES, preprocessing, adRNN and modulation (see Fig. \ref{figBD}). ES is the same as for the main track. Preprocessing component normalizes time series sequences in the same way as its main track counterpart. It uses equations \eqref{eqxt}, \eqref{eqxt2} and \eqref{eqxpt} for this. Then, it constructs the input pattern for RNN, which has the form of \eqref{eqxp} excluding modulated context vector $\textbf{r}'_t$.

Context adRNN has the same architecture as the main adRNN except that: (i) inputs do not include $\textbf{r}'_t$, (ii) exogenous vector is not modulated (no $\textbf{p}_{t})$, and (iii) 
the output is composed of $u$-component context vector $\textbf{r}_t$ and correction for ES, $\Delta\alpha_t$.

To adjust the context vector to the individual price series forecasted by the main track, this vector is element-wise multiplied by the $u$-component modulation vectors, $\textbf{g}^{(j)}, j=1, ..., J$, which are learned for each of the $J$ series from the main batch.
Modulated context vector is of the form:

\begin{equation}
\textbf{r'}_t^{(j)} = \textbf{r}_t   \otimes  \textbf{g}^{(j)}
\label{eqg}
\end{equation}

Modulation vectors $\textbf{g}^{(j)}$ as well as vectors $\textbf{p}^{(j)}$ are updated by the overall optimization procedure and have a static nature, i.e. they does not change while stepping through the batch time steps.

\section{Experimental Study}

In this section, we apply our proposed model to forecast prices of 15 cryptocurrencies and compare its performance with that of comparative models including both statistical and ML ones. 
Although cES-adRNN is able to predict both point forecasts and predictive intervals, in this study we do not explore the model's capabilities to create probabilistic forecasts, focusing on point forecasts. The model was implemented in Python using PyTorch. It was run on an eight-core CPU (AMD Ryzen 7 1700, 3.0 GHz, 32 GB RAM).

During the model development and hyperparameter search, the data used was from the period preceding the one-year test period, which spanned from 1 June 2021 to 31 May 2022. Then, using the best hyperparameters, a final training was conducted and the model was tested on the data from the one-year test period. The training procedure generally followed \cite{Smy23}, but there a few changes, necessitated by a small number of series: 
\begin{itemize}
  \item batch size was changed once only, from 2 to 4, in epoch 6,
  \item although the batch size could not be increased much, but a similar effect of obtaining more smooth and higher quality gradient was achieved by increasing number of training steps applied to each batch, starting from 15, and doubling it in epoch 2, tripling to 45 in epoch 3, increasing to 60 in epoch 4, and to 75 in epoch 5,
  \item additionally, in \cite{Smy23}, the epoch was defined as executing 2000-2500 updates; here due to very small dataset size, the overtraining was starting much earlier, forcing us to reduce the number of updates per epoch to 300-500 for horizon of 1, 150-200 for horizon of 7, and just 50-70 for horizon of 28,
   \item  we used the ensemble size of 30.
\end{itemize}

As the performance metrics, the following measures were used:
MAPE - mean absolute percentage error, RMSE - root mean square error, MPE - mean absolute percentage error, and StdPE - standard deviation of percentage error.  

The models used for comparison were:

\begin{itemize}
\item Naive -- naive model: the forecasted price for day $i$ is the same as the price for day $i-h$, where $h=1, 7$ or $28$ is the forecast horizon. 
\item ARIMA -- autoregressive integrated moving average model \cite{Dud15},
\item ES -- exponential smoothing model \cite{Dud15},
\item FNM -- fuzzy neighbourhood model \cite{Dud15}
\item MLP -- perceptron with a single hidden layer and sigmoid nonlinearities \cite{Dud16a},
\item DeepAR -- autoregressive RNN model for probabilistic forecasting \cite{Sal20},
\item N-BEATS -- deep NN with hierarchical doubly residual topology \cite{Ore20},
\item Transformer -- transformer NN with attention mechanism \cite{Vas17},
\item WaveNet -- autoregressive deep NN model combining causal filters with dilated convolutions \cite{Oor16}.
\end{itemize}

We used R implementations of ARIMA and ES (package \texttt{forecast}), and GluonTS implementations of MLP, DeepAR, N-BEATS, Transformer and WaveNet \cite{Ale20}. A Naive model and FNM were implemented in Matlab. 

Our experimentation involved two paths. The first path was to predict cryptocurrency prices solely based on their historical values, without using exogenous variables. The second path incorporated time series of both prices and exogenous variables as input.
Tables \ref{t1} and \ref{t2} show the results of both experimentation paths. The models with exogenous variables are marked with the "+" symbol (note that some of the models are not able to incorporate exogenous data).

The superiority of our proposed model over other models is clearly demonstrated by the results presented in Tables \ref{t1} and \ref{t2}: our model fed by exogenous variables, cES-adRNN+, produced the most accurate forecasts for all horizons. The differences in MAPE between cES-adRNN variants without and with exogenous inputs were: 3.0\% for $h=1$, 3.8\% for $h=7$, and 7.2\% for $h=28$. Note that in case of other models, introducing exogenous variables not always lead to error decreasing (this is especially evident for $h=1$). For a horizon of 1, it is challenging for forecasting models to outperform the Naive model. However, our proposed model is the only one that generated errors lower than those of the Naive model. For longer horizons, the Naive model produced forecasts with much greater errors than those of other models.

The cES-adRNN+ model shows the lowest StdPE values, indicating that its predictions are less dispersed compared to the baseline models. Additionally, the MPE values for cES-adRNN+ are also among the lowest. MPE reflects a forecast bias. While our model can reduce bias by selecting an appropriate quantile order for the loss function, it's important to note that bias reduction may negatively impact forecast error, which is our primary quality measure. As a result, we did not further reduce bias in our model.

To further evaluate the accuracy of the models, we conducted a pairwise one-sided Giacomini-White test for conditional predictive ability for each cryptocurrency and forecast horizon \cite{Gia06}. Table \ref{t3} shows the results: GWtest value, i.e. the percentage of cases where a given model outperformed other models in terms of MAPE at a significance level of $\alpha=0.05$. For instance, a GWtest value of 95.1 was obtained for $h=28$ by cES-adRNN+, indicating that this model had a significantly lower MAPE than other models in 95.1\% of pairwise comparisons. For horizon of 1, our model demonstrates a highest value of GWtest among all models (57.3), but note that the second highest value is for the Naive model (45.8). For longer horizons, GWtest increased for cES-adRNN+ to 89.8 for $h=7$, and 95.1 for $h=28$.

\begin{figure}[h]
\centering
    \includegraphics[width=0.30\textwidth]{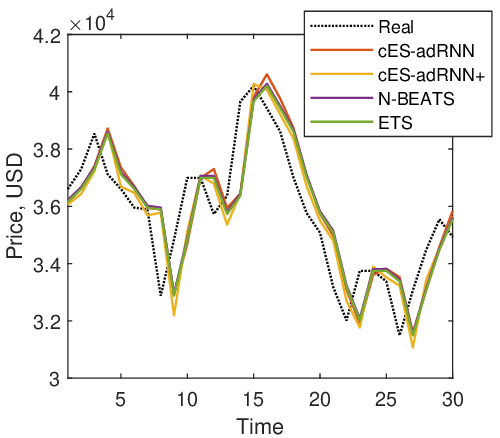}
    \includegraphics[width=0.30\textwidth]{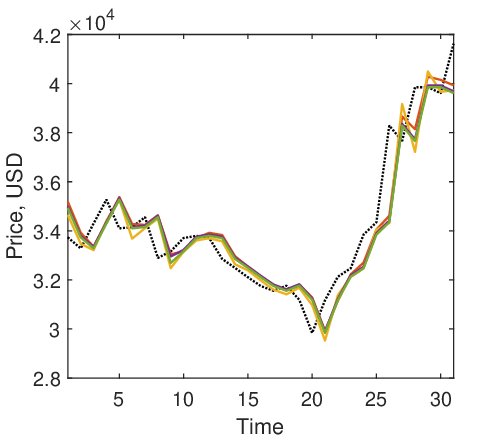}
    \includegraphics[width=0.30\textwidth]{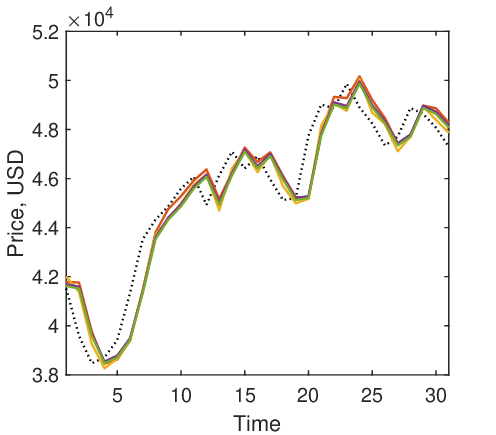}
    \includegraphics[width=0.30\textwidth]{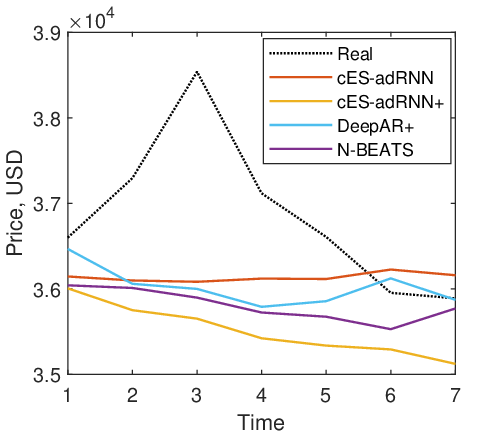}
    \includegraphics[width=0.30\textwidth]{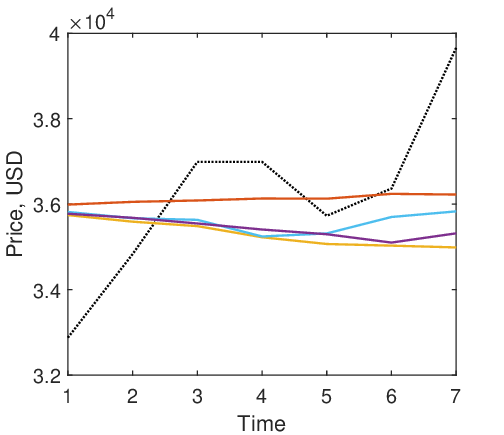}
    \includegraphics[width=0.30\textwidth]{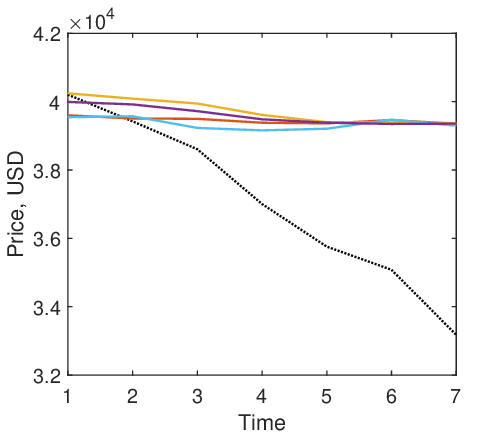}
    \includegraphics[width=0.30\textwidth]{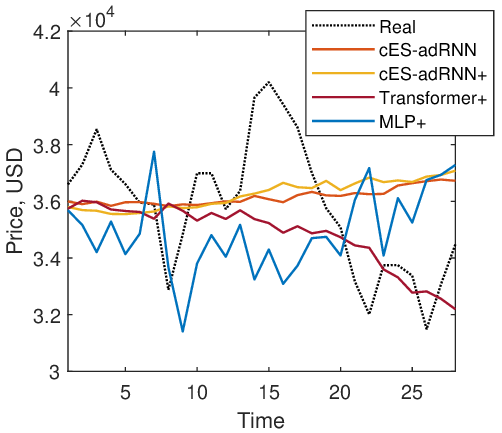}
    \includegraphics[width=0.30\textwidth]{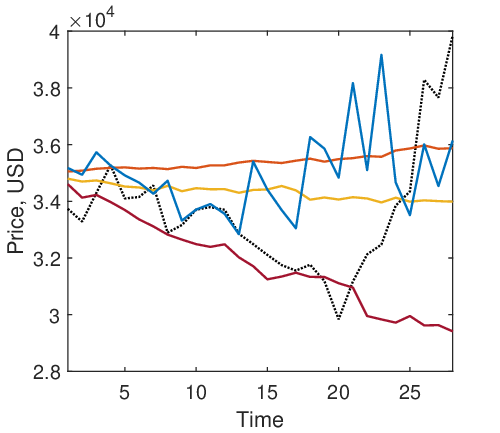}
    \includegraphics[width=0.30\textwidth]{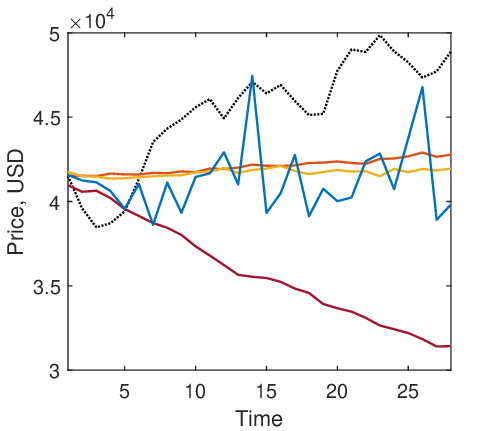} 
\caption{Examples of forecasts for BTC: top panel - $h=1$, middle panel - $h=7$, bottom panel - $h=28$.} 
\label{figPr}
\end{figure}

Figure \ref{figPr} presents examples of BTC price forecasts generated by the most accurate models. The top panel displays one-day ahead forecasts for the first three months of the test period, revealing a characteristic inertia or lag in the predictions. The middle panel shows 7-day ahead forecasts for the first three weeks of the test period, while the bottom panel displays 28-day ahead forecasts for the first three months. Notably, the models are unable to accurately predict BTC price fluctuations for longer horizons due to insufficient information in the input variables.

 \begin{table}[h]
  \centering
  \caption{Forecasting metrics for models without exogenous variables (univariate case).}
  \scriptsize
    \begin{tabular}{|l|r|r|r|r|r|r|r|r|r|r|r|r|} 
    \hline
    \multicolumn{1}{|l|}{} & \multicolumn{4}{c|}{Horizon 1} & \multicolumn{4}{c|}{Horizon 7} & \multicolumn{4}{c|}{Horizon 28} \\    
    \cline{2-13}
%Model &  \mfs MAPE& \mfs RMSE& \mfs MPE& \mfs StdPE& \mfs MAPE& \mfs RMSE& \mfs MPE& \mfs StdPE& \mfs MAPE& \mfs RMSE& \mfs MPE& \mfs StdPE \\
Model &   MAPE&  RMSE&  MPE&  StdPE&  MAPE&  RMSE&  MPE&  StdPE&  MAPE&  RMSE&  MPE&  StdPE \\

 \hline
    Naive & \textbf{3.25}  & \textbf{70.6}  & { } 0.36  & \textbf{4.68}  & 10.63 & 256.0 & { } 2.86  & 11.52 & 22.43 & 567.3 & { } 9.66  & 20.87 \\
    ARIMA & 3.38  & 73.7  & \textbf{0.22}  & 4.82  & 7.92  & 194.4 & 1.23  & 9.72  & 16.45 & 418.5 & 3.99  & 18.32 \\
    ES   & 3.29  & 71.1  & 0.33  & 4.80  & 7.53  & 181.3 & 1.56  & 9.84  & 15.60 & 386.2 & 5.27  & 21.63 \\
    FNM   & 5.48  & 101.8 & 0.97  & 7.49  & 10.18 & 249.0 & 3.35  & 11.79 & 20.79 & 530.9 & 9.51  & 23.75 \\
    MLP   & 3.38  & 75.1  & 0.86  & 4.75  & 7.87  & 198.7 & 1.80  & 9.21  & 15.97 & 462.1 & \textbf{2.05}  & 15.89 \\
    DeepAR & 3.30  & 72.1  & 0.33  & 4.73  & 7.71  & 191.5 & 2.41  & 9.14  & 16.68 & \textbf{381.4} & 8.73  & 16.02 \\
    N-BEATS & 3.27  & 71.0  & 0.48  & 4.69  & \textbf{7.33}  &\textbf{ 179.8} & \textbf{0.06}  & \textbf{8.81}  & 16.48 & 382.6 & 5.79  & 16.46 \\
    Transformer & 5.44  & 157.7 & 4.68  & 4.95  & 8.09  & 188.5 & 3.09  & 9.42  & 17.95 & 647.1 & 3.42  & 18.22 \\
    WaveNet & 10.23 & 188.9 & 4.84  & 15.89 & 8.10  & 191.3 & 3.48  & 9.75  & 18.18 & 403.0 & 10.56 & 17.94 \\
    cES-adRNN & 3.29  & 72.3  & 0.39  & 4.75  & 7.40  & 184.2 & 1.58  & 8.99  & \textbf{15.05} & 391.7 & 5.40  & \textbf{15.74} \\
    \hline  
    \end{tabular}%    
  \label{t1}%
\end{table}%
\begin{table}[h]
  \centering
  \setlength{\tabcolsep}{1.pt}
  \caption{Forecasting metrics for models with exogenous variables (multivariate case).}
  \scriptsize
    \begin{tabular}{|l|r|r|r|r|r|r|r|r|r|r|r|r|} 
    \hline
    \multicolumn{1}{|l|}{} & \multicolumn{4}{c|}{Horizon 1} & \multicolumn{4}{c|}{Horizon 7} & \multicolumn{4}{c|}{Horizon 28} \\    
    \cline{2-13}
%Model &  \mfs MAPE& \mfs RMSE& \mfs MPE& \mfs StdPE& \mfs MAPE& \mfs RMSE& \mfs MPE& \mfs StdPE& \mfs MAPE& \mfs RMSE& \mfs MPE& \mfs StdPE \\
Model &   MAPE&  RMSE&  MPE&  StdPE&  MAPE&  RMSE&  MPE&  StdPE&  MAPE&  RMSE&  MPE&  StdPE \\
 \hline
   
  MLP+  & 3.33  & 75.4  & 0.31  & 4.74  & 7.74  & 201.1 & 1.20  & 9.06  & 15.54 & 430.7 & 4.64  & 15.88 \\
    DeepAR+ & 3.31  & 71.8  & 0.42  & 4.72  & 7.27  & 181.0 & \textbf{-0.27} & \textbf{8.59}  & 15.95 & 432.8 & 8.04  & 15.82 \\
    N-BEATS+ & 3.35  & 70.7  & 0.69  & 4.70  & 7.35  & 179.8 & 0.81  & 8.85  & 16.87 & 375.5 & 6.24  & 16.29 \\
    Transformer+ & 5.91  & 173.6 & 5.28  & 4.96  & 7.54  & 188.6 & 0.61  & 9.12  & 14.50 & 468.1 & \textbf{0.26} & 15.35 \\
    WaveNet+ & 4.01  & 86.4  & 2.10  & 6.56  & 7.86  & 187.0 & 3.00  & 9.48  & 19.29 & 572.0 & 12.98 & 17.50 \\
    cES-adRNN+ & \textbf{3.19}  & \textbf{68.9}  & \textbf{0.21}  & \textbf{4.58}  & \textbf{7.12}  & \textbf{179.6} & 0.55  & 8.67  & \textbf{13.97} & \textbf{374.3} & 2.08  & \textbf{15.11} \\

    \hline  
    \end{tabular}%    
  \label{t2}%
\end{table}%

\begin{table}[h]
  \centering
    \setlength{\tabcolsep}{2.4pt}
  \caption{GWtest metric.}
  \scriptsize
   % \begin{tabular}{|l|rrrrrrrrrrrrrrrr|}
    \begin{tabular}{|l|r|r|r|r|r|r|r|r|r|r|r|r|r|r|r|r|}
    \hline
 & \rotatebox[origin=l]{90}{Naive}& \rotatebox[origin=l]{90}{ARIMA}& \rotatebox[origin=l]{90}{ETS}& \rotatebox[origin=l]{90}{FNM}& \rotatebox[origin=l]{90}{MLP}&  \rotatebox[origin=l]{90}{DeepAR}& \rotatebox[origin=l]{90}{N-BEATS}&  \rotatebox[origin=l]{90}{Transformer}&  \rotatebox[origin=l]{90}{WaveNet}& \rotatebox[origin=l]{90}{cES-adRNN}& 
 \rotatebox[origin=l]{90}{MLP+}&
 \rotatebox[origin=l]{90}{DeepAR+}&
 \rotatebox[origin=l]{90}{N-BEATS+}&
 \rotatebox[origin=l]{90}{Transformer+}&
 \rotatebox[origin=l]{90}{WaveNet+}& \rotatebox[origin=l]{90}{cES-adRNN+}\\

    \hline
         Horizon 1  & 45.8  & 29.8  & 32.0  & 15.6  & 30.2  & 15.6  & 31.1  & 8.4   & 31.1  & 0.0   & 31.6  & 25.8  & 39.1  & 31.1  & 36.4  & \textbf{57.3} \\
    Horizon 7  & 2.2   & 32.4  & 53.3  & 5.3   & 20.4  & 27.1  & 31.6  & 52.4  & 40.0  & 30.7  & 76.4  & 44.0  & 77.3  & 67.6  & 78.7  & \textbf{89.8} \\
    Horizon 28  & 3.6   & 44.0  & 64.4  & 10.7  & 43.6  & 25.3  & 56.9  & 72.4  & 42.7  & 22.7  & 57.8  & 15.6  & 75.1  & 69.3  & 83.1  & \textbf{95.1} \\
    \hline
    \end{tabular}%
  \label{t3}%
\end{table}%

The model training process took several hours on a desktop-class computer, utilizing only CPUs (without GPUs) running in parallel with all available cores. The final forecast was then aggregated after the training. The forecasting process is expected to be much faster, as it does not involve gradient calculations. In a real-life scenario, we would expect daily forecasts based on saved NN weights, which could be generated in less than a minute. The retraining process could be performed less frequently, for example, on a monthly basis.

\section{Conclusions}

Cryptocurrencies are notoriously volatile and their prices are influenced by a range of factors, such as market sentiment, regulatory changes, and technological advancements. Moreover, the lack of discernible patterns in the price data and its erratic fluctuations make accurately predicting cryptocurrency prices a daunting task. Despite these challenges, our study employs state-of-the-art ML model, cES-sdRNN, that incorporates various mechanisms and procedures to improve forecasting efficiency, such as a hybrid architecture, context track, recurrent cells with dilation and attention mechanisms, dynamic ES model, cross-learning, quantile loss function, ensembling, and overfitting prevention mechanisms. Our proposed model, which uses exogenous variables as input, outperformed all comparative models, achieving the highest accuracy across all forecast horizons.

%Our research findings suggest that the presented models are not effective and trustworthy predictors of cryptocurrency prices. This outcome could be attributed to the complexity of the problem, which advanced deep learning techniques such as LSTM and CNNs are unable to efficiently address. Similar 
%Other researchers come to similar conclusions, see \cite{Pin20}. 

%"Also, based on our experimental results and investigation regarding to our research questions about cryptocurrency price problem, we conclude that cryptocurrency prices follow almost a random walk process while few hidden patterns may probably exist in, where an intelligent framework has to identify them in order for a prediction model to make accurate and reliable forecasts. Therefore, new sophisticated algorithmic methods, alternative approaches, new validation metrics should be explored."

%\bibitem{Pin20}
%Pintelas, E., Livieris, I.E., Stavroyiannis, S., Kotsilieris, T., Pintelas, P. (2020). Investigating the Problem of Cryptocurrency Price Prediction: A Deep Learning Approach. In: Maglogiannis, I., Iliadis, L., Pimenidis, E. (eds) Artificial Intelligence Applications and Innovations. AIAI 2020. IFIP Advances in Information and Communication Technology, vol 584. Springer, Cham. %https://doi.org/10.1007/978-3-030-49186-4_9


\end{document}